\documentclass{article}
\usepackage[latin9]{inputenc}
\usepackage{array}
\usepackage{float}
\usepackage{multirow}
\usepackage{graphicx}
\usepackage[authoryear]{natbib}
\usepackage[unicode=true,
 bookmarks=false,
 breaklinks=false,pdfborder={0 0 1},backref=false,colorlinks=false]
 {hyperref}

\makeatletter

\providecommand{\tabularnewline}{\\}
\newcommand{\lyxdot}{.}

\usepackage{iclr2017_conference}
\usepackage{times}
\usepackage{url}
\usepackage{amsfonts}
\usepackage{enumitem}
\setlist{nolistsep}
\date{}

\iclrfinalcopy 

\@ifundefined{showcaptionsetup}{}{%
 \PassOptionsToPackage{caption=false}{subfig}}
\usepackage{subfig}
\makeatother

\begin{document}

\title{Learning to Generate Samples from Noise through Infusion Training}

\author{Florian Bordes, Sina Honari, Pascal Vincent\thanks{Associate Fellow, Canadian Institute For Advanced Research (CIFAR)}\\
Montreal Institute for Learning Algorithms (MILA)\\
Département d'Informatique et de Recherche Opérationnelle\\
Université de Montréal\\
Montréal, Québec, Canada\\
\texttt{\{firstname.lastname@umontreal.ca\}}}

\maketitle
\maketitle
\begin{abstract}
In this work, we investigate a novel training procedure to learn a
generative model as the transition operator of a Markov chain, such
that, when applied repeatedly on an unstructured random noise sample,
it will denoise it into a sample that matches the target distribution
from the training set. The novel training procedure to learn this
progressive denoising operation involves sampling from a slightly
different chain than the model chain used for generation in the absence
of a denoising target. In the training chain we infuse information
from the training target example that we would like the chains to
reach with a high probability. The thus learned transition operator
is able to produce quality and varied samples in a small number of
steps. Experiments show competitive results compared to the samples
generated with a basic Generative Adversarial Net.

\end{abstract}

\section{Introduction and motivation}

To go beyond the relatively simpler tasks of classification and regression,
advancing our ability to learn good generative models of high-dimensional
data appears essential. There are many scenarios where one needs to
efficiently produce good high-dimensional outputs where output dimensions
have unknown intricate statistical dependencies: from generating realistic
images, segmentations, text, speech, keypoint or joint positions,
etc..., possibly as an answer to the same, other, or multiple input
modalities. These are typically cases where there is not just one
right answer but a variety of equally valid ones following a non-trivial
and unknown distribution. A fundamental ingredient for such scenarios
is thus the ability to learn a good generative model from data, one
from which we can subsequently efficiently generate varied samples
of high quality. 

Many approaches for learning to generate high dimensional samples
have been and are still actively being investigated. These approaches
can be roughly classified under the following broad categories:
\begin{itemize}
\item Ordered visible dimension sampling \citep{OordKK16-pixelRNN,larochelle2011neural}.
In this type of auto-regressive approach, output dimensions (or groups
of conditionally independent dimensions) are given an arbitrary fixed
ordering, and each is sampled conditionally on the previous sampled
ones. This strategy is often implemented using a recurrent network
(LSTM or GRU). Desirable properties of this type of strategy are that
the exact log likelihood can usually be computed tractably, and sampling
is exact. Undesirable properties follow from the forced ordering,
whose arbitrariness feels unsatisfactory especially for domains that
do not have a natural ordering (e.g. images), and imposes for high-dimensional
output a \emph{long} \emph{sequential} generation that can be slow. 
\item Undirected graphical models with \emph{multiple} layers of \emph{latent}
variables. These make inference, and thus learning, particularly hard
and tend to be costly to sample from \citep{salakhutdinov2009deep}.
\item Directed graphical models trained as variational autoencoders (VAE)
\citep{kingma2014auto,DBLP:conf/icml/RezendeMW14}
\item Adversarially-trained generative networks. (GAN)\citep{Goodfellow-et-al-NIPS2014-small}
\item Stochastic neural networks, i.e. networks with stochastic neurons,
trained by an adapted form of stochastic backpropagation
\item Generative uses of denoising autoencoders \citep{vincent2010stacked}
and their generalization as Generative Stochastic Networks \citep{alain-al-iai2016-gsns}
\item Inverting a non-equilibrium thermodynamic slow \emph{diffusion} process
\citep{Sohl-Dickstein-et-al-ICML2015}
\item Continuous transformation of a distribution by invertible functions
(\citet{dinh2014nice}, also used for variational inference in \citet{rezende2015variational})
\end{itemize}
Several of these approaches are based on maximizing an explicit or
implicit model log-likelihood or a lower bound of its log-likelihood,
but some successful ones are not e.g. GANs. The approach we propose
here is based on the notion of ``denoising'' and thus takes its
root in denoising autoencoders and the GSN type of approaches. It
is also highly related to the non-equilibrium thermodynamics inverse
diffusion approach of \citet{Sohl-Dickstein-et-al-ICML2015}. One
key aspect that distinguishes these types of methods from others listed
above is that sample generation is achieved thanks to a learned stochastic
mapping from input space to input space, rather than from a latent-space
to input-space. 

Specifically, in the present work, we propose to learn to generate
high quality samples through a process of \emph{progressive}, \emph{stochastic,
denoising}, starting from a simple initial ``noise'' sample generated
in input space from a simple factorial distribution i.e. one that
does not take into account any dependency or structure between dimensions.
This, in effect, amounts to learning the transition operator of a
Markov chain operating on input space. Starting from such an initial
``noise'' input, and repeatedly applying the operator for a small
fixed number $T$ of steps, we aim to obtain a high quality resulting
sample, effectively modeling the training data distribution. Our training
procedure uses a novel ``target-infusion'' technique, designed to
slightly bias model sampling to move towards a specific data point
during training, and thus provide inputs to denoise which are likely
under the model's sample generation paths. By contrast with \citet{Sohl-Dickstein-et-al-ICML2015}
which consists in inverting a slow and fixed diffusion process, our
infusion chains make a few large jumps and follow the model distribution
as the learning progresses. 

The rest of this paper is structured as follows: Section 2 formally
defines the model and training procedure. Section 3 discusses and
contrasts our approach with the most related methods from the literature.
Section 4 presents experiments that validate the approach. Section
5 concludes and proposes future work directions.

\section{Proposed approach}

\subsection{Setup}

We are given a finite data set $D$ containing $n$ points in $\mathbb{R}^{d}$,
supposed drawn i.i.d from an unknown distribution $q^{*}$. The data
set $D$ is supposed split into training, validation and test subsets
$D_{\mathrm{train}}$, $D_{\mathrm{valid}}$, $D_{\mathrm{test}}$.
We will denote $q_{\mathrm{train}}^{*}$ the \emph{empirical distribution}
associated to the training set, and use $\mathbf{x}$ to denote observed
samples from the data set. We are interested in learning the parameters
of a generative model $p$ conceived as a Markov Chain from which
we can efficiently sample. Note that we are interested in learning
an operator that will display fast \emph{``burn-in''} from the initial
factorial ``noise'' distribution, but beyond the initial $T$ steps
we are not concerned about potential slow mixing or being stuck. We
will first describe the sampling procedure used to sample from a trained
model, before explaining our training procedure.

\subsection{Generative model sampling procedure}

\begin{figure}
\includegraphics[width=1\columnwidth]{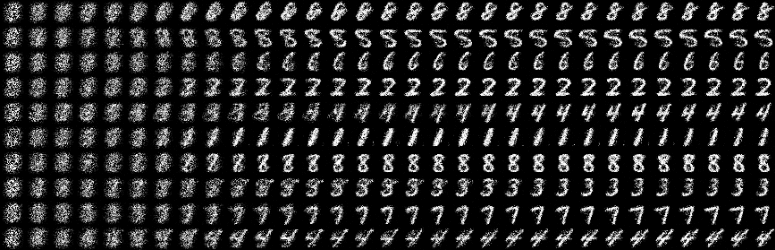}

\protect\caption{\label{fig:model-sampling-chain}The \emph{model sampling chain}.
Each row shows a sample from $p(\mathbf{z}^{(0)},\ldots,\mathbf{z}^{(T)})$
for a model that has been trained on MNIST digits. We see how the
learned Markov transition operator progressively denoises an initial
unstructured noise sample. We can also see that there remains ambiguity
in the early steps as to what digit this could become. This ambiguity
gets resolved only in later steps. Even after a few initial steps,
stochasticity could have made a chain move to a different final digit
shape.}

\end{figure}

The generative model $p$ is \emph{defined} as the following sampling
procedure:
\begin{itemize}
\item Using a simple factorial distribution $p^{(0)}(\mathbf{z}^{(0)})$,
draw an initial sample $\mathbf{z}^{(0)}\sim p^{(0)}$, where $\mathbf{z}^{(0)}\in\mathbb{R}^{d}$.
Since $p^{(0)}$ is factorial, the $d$ components of $\mathbf{z}^{(0)}$
are independent: $p^{0}$ cannot model any dependency structure. $\mathbf{z}^{(0)}$
can be pictured as essentially unstructured random noise.
\item Repeatedly apply $T$ times a stochastic transition operator $p^{(t)}(\mathbf{z}^{(t)}|\mathbf{z}^{(t-1)})$,
yielding a more ``denoised'' sample $\mathbf{z}^{(t)}\sim p^{(t)}(\mathbf{z}^{(t)}|\mathbf{z}^{(t-1)})$,
where all $\mathbf{z}^{(t)}\in\mathbb{R}^{d}$.
\item Output $\mathbf{z}^{(T)}$ as the final generated sample. Our generative
model distribution is thus $p(\mathbf{z}^{(T)})$, the marginal associated
to joint $p(\mathbf{z}^{(0)},\ldots,\mathbf{z}^{(T)})=p^{(0)}(\mathbf{z}^{(0)})~\left(\prod_{t=1}^{T}p^{(t)}(\mathbf{z}^{(t)}|\mathbf{z}^{(t-1)})\right)$.
\end{itemize}
In summary, samples from model $p$ are generated, starting with an
initial sample from a simple distribution $p^{(0)}$, by taking the
$T^{\mathrm{th}}$sample along Markov chain $\mathbf{z}^{(0)}\rightarrow\mathbf{z}^{(1)}\rightarrow\mathbf{z}^{(2)}\rightarrow\ldots\rightarrow\mathbf{z}^{(T)}$
whose transition operator is $p^{(t)}(\mathbf{z}^{(t)}|\mathbf{z}^{(t-1)})$.
We will call this chain the \emph{model} \emph{sampling chain}. Figure
\ref{fig:model-sampling-chain} illustrates this sampling procedure
using a model (i.e. transition operator) that was trained on MNIST.
Note that we impose no formal requirement that the chain converges
to a stationary distribution, as we simply read-out $\mathbf{z}^{(T)}$
as the samples from our model $p$. The chain also needs not be time-homogeneous,
as highlighted by notation $p^{(t)}$ for the transitions. 

The set of parameters $\theta$ of model $p$ comprise the parameters
of $p^{(0)}$ and the parameters of transition operator $p^{(t)}(\mathbf{z}^{(t)}|\mathbf{z}^{(t-1)})$.
For tractability, learnability, and efficient sampling, these distributions
will be chosen factorial, i.e. $p^{(0)}(\mathbf{z}^{(0)})=\prod_{i=1}^{d}~p_{i}^{(0)}(\mathbf{z}_{i}^{(0)})$
and $p^{(t)}(\mathbf{z}^{(t)}|\mathbf{z}^{(t-1)})=\prod_{i=1}^{d}~p_{i}^{(t)}(\mathbf{z}_{i}^{(t)}|\mathbf{z}^{(t-1)})$.
Note that the conditional distribution of an individual component
$i$, $p_{i}^{(t)}(\mathbf{z}_{i}^{(t)}|\mathbf{z}^{(t-1)})$ may
however be multimodal, e.g. a mixture in which case $p^{(t)}(\mathbf{z}^{(t)}|\mathbf{z}^{(t-1)})$
would be a product of independent mixtures (conditioned on $\mathbf{z}^{(t-1)}$),
one per dimension. In our experiments, we will take the $p^{(t)}(\mathbf{z}^{(t)}|\mathbf{z}^{(t-1)})$
to be simple diagonal Gaussian yielding a Deep Latent Gaussian Model
(DLGM) as in \citet{DBLP:conf/icml/RezendeMW14}.

\subsection{Infusion training procedure}

We want to train the parameters of model $p$ such that samples from
$D_{\mathrm{train}}$ are likely of being generated under the \emph{model
sampling chain}. Let $\theta^{(0)}$ be the parameters of $p^{(0)}$
and let $\theta^{(t)}$ be the parameters of $p^{(t)}(\mathbf{z}^{(t)}|\mathbf{z}^{(t-1)})$.
Note that parameters $\theta^{(t)}$ for $t>0$ can straightforwardly
be shared across time steps, which we will be doing in practice. Having
committed to using (conditionally) \emph{factorial} distributions
for our $p^{(0)}(\mathbf{z}^{(0)})$ and $p^{(t)}(\mathbf{z}^{(t)}|\mathbf{z}^{(t-1)})$,
that are both easy to learn and cheap to sample from, let us first
consider the following greedy stagewise procedure. We can easily learn
$p_{i}^{(0)}(\mathbf{z}^{(0)})$ to model the marginal distribution
of each component $\mathbf{x}_{i}$ of the input, by training it by
gradient descent on a maximum likelihood objective, i.e. 
\begin{equation}
\theta^{(0)}=\arg\max_{\theta}~\mathbb{E}_{\mathbf{x}\sim q_{\mathrm{train}}^{*}}\left[\log p^{(0)}(\mathbf{x};\theta)\right]\label{eq:train-p0}
\end{equation}

This gives us a first, very crude unstructured (factorial) model of
$q^{*}$.

Having learned this $p^{(0)}$, we might be tempted to then greedily
learn the next stage $p^{(1)}$ of the chain in a similar fashion,
after drawing samples $\mathbf{z}^{(0)}\sim p^{(0)}$ in an attempt
to learn to ``denoise'' the sampled $\mathbf{z}^{(0)}$ into $\mathbf{x}$.
Yet the corresponding following training objective $\theta^{(1)}=\arg\max_{\theta}~\mathbb{E}_{\mathbf{x}\sim q_{\mathrm{train}}^{*},\mathbf{z}^{(0)}\sim p^{(0)}}\left[\log p^{(1)}(\mathbf{x}|\mathbf{z}^{(0)};\theta)\right]$
makes no sense: $\mathbf{x}$ and $\mathbf{z}^{(0)}$ are sampled
independently of each other so $\mathbf{z}^{(0)}$ contains no information
about $\mathbf{x}$, hence $p^{(1)}(\mathbf{x}|\mathbf{z}^{(0)})=p^{(1)}(\mathbf{x})$.
So maximizing this second objective becomes essentially the same as
what we did when learning $p^{(0)}$. We would learn nothing more.
It is essential, if we hope to learn a useful conditional distribution
$p^{(1)}(\mathbf{\mathbf{\mathbf{x}}}|\mathbf{z}^{(0)})$ that it
be trained on \emph{particular} $\mathbf{z}^{(0)}$ containing some
information about $\mathbf{x}$. In other words, we should not take
our training inputs to be samples from $p^{(0)}$ but from a slightly
different distribution, biased towards containing some information
about $\mathbf{x}$. Let us call it $q^{(0)}(\mathbf{z}^{(0)}|\mathbf{x})$.
A natural choice for it, if it were possible, would be to take $q^{(0)}(\mathbf{z}^{(0)}|\mathbf{x})=p(\mathbf{z}^{(0)}|\mathbf{z}^{(T)}=\mathbf{x})$
but this is an intractable inference, as all intermediate $\mathbf{z}^{(t)}$
between $\mathbf{z}^{(0)}$ and $\mathbf{z}^{(T)}$ are effectively
latent states that we would need to marginalize over. Using a workaround
such as a variational or MCMC approach would be a usual fallback.
Instead, let us focus on our initial intent of guiding a progressive
stochastic denoising, and think if we can come up with a different
way to construct $q^{(0)}(\mathbf{z}^{(0)}|\mathbf{x})$ and similarly
for the next steps $q_{i}^{(t)}(\tilde{\mathbf{z}}_{i}^{(t)}|\tilde{\mathbf{z}}^{(t-1)},\mathbf{x})$.

Eventually, we expect a sequence of samples from Markov chain $p$
to move from initial ``noise'' towards a specific example $\mathbf{x}$
from the training set rather than another one, primarily if a sample
along the chain ``resembles'' $\mathbf{x}$ to some degree. This
means that the transition operator should learn to pick up a minor
resemblance with an $\mathbf{x}$ in order to transition to something
likely to be even more similar to $\mathbf{x}$. In other words, we
expect samples along a chain leading to $\mathbf{x}$ to both have
high probability under the transition operator of the chain $p^{(t)}(\mathbf{z}^{(t)}|\mathbf{z}^{(t-1)})$,
\emph{and} to have some form of at least partial ``resemblance''
with $\mathbf{x}$ likely to increase as we progress along the chain.
One highly inefficient way to emulate such a chain of samples would
be, for teach step $t$, to sample many candidate samples from the
transition operator (a conditionally factorial distribution) until
we generate one that has some minimal ``resemblance'' to $\mathbf{x}$
(e.g. for a discrete space, this resemblance measure could be based
on their Hamming distance). A qualitatively similar result can be
obtained at a negligible cost by sampling from a factorial distribution
that is very close to the one given by the transition operator, but
very slightly biased towards producing something closer to $\mathbf{x}$.
Specifically, we can ``infuse'' a little of $\mathbf{x}$ into our
sample by choosing for each input dimension, whether we sample it
from the distribution given for that dimension by the transition operator,
or whether, with a small probability, we take the value of that dimension
from $\mathbf{x}$. Samples from this biased chain, in which we slightly
``infuse'' $\mathbf{x}$, will provide us with the inputs of our
input-target training pairs for the transition operator. The target
part of the training pairs is simply $\mathbf{x}$.

\subsubsection{The infusion chain}

Formally we define an \emph{infusion} \emph{chain} $\widetilde{\mathbf{z}}^{(0)}\rightarrow\widetilde{\mathbf{z}}^{(1)}\rightarrow\ldots\rightarrow\widetilde{\mathbf{z}}^{(T-1)}$
whose distribution $q(\widetilde{\mathbf{z}}^{(0)},\ldots,\widetilde{\mathbf{z}}^{(T-1)}|\mathbf{x})$
will be ``close'' to the \emph{sampling chain} $\mathbf{z}^{(0)}\rightarrow\mathbf{z}^{(1)}\rightarrow\mathbf{z}^{(2)}\rightarrow\ldots\rightarrow\mathbf{z}^{(T-1)}$
of model $p$ in the sense that $q^{(t)}(\tilde{\mathbf{z}}^{(t)}|\tilde{\mathbf{z}}^{(t-1)},\mathbf{x})$
will be close to $p^{(t)}(\mathbf{z}^{(t)}|\mathbf{z}^{(t-1)})$,
but will at every step be slightly biased towards generating samples
closer to target $\mathbf{x}$, i.e. $\mathbf{x}$ gets progressively
``infused'' into the chain. This is achieved by defining $q_{i}^{(0)}(\widetilde{\mathbf{z}}_{i}^{(0)}|\mathbf{x})$
as a mixture between $p_{i}^{(0)}$ (with a large mixture weight)
and $\delta_{\mathbf{x_{i}}}$, a concentrated unimodal distribution
around $\mathbf{x}_{i}$, such as a Gaussian with small variance (with
a small mixture weight)%
\footnote{Note that $\delta_{\mathbf{x_{i}}}$ does not denote a Dirac-Delta
but a Gaussian with small sigma. %
}. Formally $q_{i}^{(0)}(\tilde{\mathbf{z}}{}_{i}^{(0)}|\mathbf{x})=(1-\alpha^{(t)})p_{i}^{(0)}(\tilde{\mathbf{z}}_{i}^{(0)})+\alpha^{(t)}\delta_{\mathbf{x_{i}}}(\tilde{\mathbf{z}}_{i}^{(0)})$,
where $1-\alpha^{(t)}$ and $\alpha^{(t)}$ are the mixture weights
\footnote{In all experiments, we use an increasing schedule $\alpha^{(t)}=\alpha^{^{(t-1)}}+\omega$
with $\alpha^{^{(0)}}$ and $\omega$ constant. This allows to build
our chain such that in the first steps, we give little information
about the target and in the last steps we give more informations about
the target. This forces the network to have less confidence (greater
incertitude) at the beginning of the chain and more confidence on
the convergence point at the end of the chain.%
}. In other words, when sampling a value for $\tilde{\mathbf{z}}{}_{i}^{(0)}$
from $q_{i}^{(0)}$ there will be a small probability $\alpha^{(0)}$
to pick value close to $\mathbf{x}_{i}$ (as sampled from $\delta_{\mathbf{x_{i}}}$)
rather than sampling the value from $p_{i}^{(0)}$. We call $\alpha^{(t)}$
the \emph{infusion rate}. We define the transition operator of the
\emph{infusion chain} similarly as: $q_{i}^{(t)}(\tilde{\mathbf{z}}_{i}^{(t)}|\tilde{\mathbf{z}}^{(t-1)},\mathbf{x})=(1-\alpha^{(t)})p_{i}^{(t)}(\tilde{\mathbf{z}}_{i}^{(t)}|\mathbf{\tilde{\mathbf{z}}}^{(t-1)})+\alpha^{(t)}\delta_{\mathbf{x}_{i}}(\tilde{\mathbf{z}}_{i}^{(t)})$.

\subsubsection{Denoising-based Infusion training procedure}

\begin{figure}
\raggedleft{}\includegraphics[width=1\columnwidth]{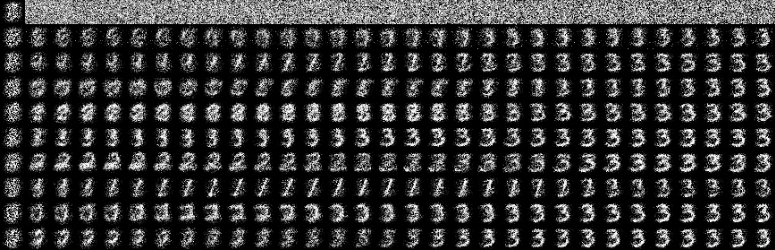}\protect\caption{\label{fig:Training-infusion-chains}Training \emph{infusion chains,}
infused with \textbf{target $\mathbf{x=}$}\protect\includegraphics[scale=0.2]{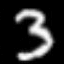}.
This figure shows the evolution of chain $q(\mathbf{z}^{(0)},\ldots,\mathbf{z}^{(30)}|\mathbf{x})$
as training on MNIST progresses. Top row is after network random weight
initialization. Second row is after 1 training epochs, third after
2 training epochs, and so on. Each of these images were at a time
provided as the input part of the (\emph{input},\textbf{ }\emph{target})
training pairs for the network. The network was trained to denoise
all of them into target 3. We see that as training progresses, the
model has learned to pick up the cues provided by target infusion,
to move towards that target. Note also that a single denoising step,
even with target infusion, is not sufficient for the network to produce
a sharp well identified digit.}
\end{figure}

For all $\mathbf{x}\in D_{\mathrm{train}}$:
\begin{itemize}
\item Sample from the \emph{infusion chain} $\mathbf{\tilde{\mathbf{z}}}=(\tilde{\mathbf{z}}^{(0)},\ldots,\tilde{\mathbf{z}}^{(T-1)})\sim q(\tilde{\mathbf{z}}^{(0)},\ldots,\tilde{\mathbf{z}}^{(T-1)}|\mathbf{x})$.
\\
precisely so: $\tilde{\mathbf{z}}_{0}\sim q^{(0)}(\tilde{\mathbf{z}}^{(0)}|\mathbf{x})$~
$\ldots$~~$\tilde{\mathbf{z}}^{(t)}\sim q^{(t)}(\tilde{\mathbf{z}}^{(t)}|\tilde{\mathbf{z}}^{(t-1)},\mathbf{x})$~~
$\ldots$
\item Perform a gradient step so that $p$ learns to ``denoise'' every
$\tilde{\mathbf{z}}^{(t)}$ into $\mathbf{x}$.\\
\[
\theta^{(t)}\leftarrow\theta^{(t)}+\eta^{(t)}\frac{\partial\log p^{(t)}(\mathbf{x}|\tilde{\mathbf{z}}^{(t-1)};\theta^{(t)})}{\partial\theta^{(t)}}
\]
 where $\eta^{(t)}$ is a scalar learning rate. %
\footnote{Since we will be sharing parameters between the $p^{(t)}$, in order
for the expected larger error gradients on the earlier transitions
not to dominate the parameter updates over the later transitions we
used an increasing schedule $\eta^{(t)}=\eta_{0}\frac{t}{T}$ for
$t\in\{1,\ldots,T\}$.%
}
\end{itemize}
As illustrated in Figure \ref{fig:Training-infusion-chains}, the
distribution of samples from the infusion chain evolves as training
progresses, since this chain remains close to the model sampling chain.

\paragraph{}

\subsection{Stochastic log likelihood estimation\label{sub:Stochastic-log-likelihood}}

The exact log-likelihood of the generative model implied by our model
$p$ is intractable. The log-probability of an example $\mathbf{x}$
can however be expressed using proposal distribution $q$ as:

\begin{equation}
\log p(\mathbf{x})=\log\mathbb{E}_{q(\widetilde{\mathbf{z}}|\mathbf{x})}\left[\frac{p(\mathbf{\tilde{\mathbf{z}}},\mathbf{x})}{q(\widetilde{\mathbf{z}}|\mathbf{x})}\right]\label{eq:is-loglikelihood}
\end{equation}

Using Jensen's inequality we can thus derive the following lower bound:

\begin{eqnarray}
\log p(\mathbf{x}) & \geq & \mathbb{E}_{q(\widetilde{\mathbf{z}}|\mathbf{x})}\left[\log p(\mathbf{\tilde{\mathbf{z}}},\mathbf{x})-\log q(\widetilde{\mathbf{z}}|\mathbf{x})\right]\label{eq:lower-bound}
\end{eqnarray}

where $\log p(\mathbf{\tilde{\mathbf{z}}},\mathbf{x})=\log p^{(0)}(\tilde{\mathbf{z}}^{(0)})+\left(\sum_{t=1}^{T-1}\log p^{(t)}(\tilde{\mathbf{z}}^{(t)}|\tilde{\mathbf{z}}^{(t-1)})\right)+\log p^{(T)}(\mathbf{x}|\tilde{\mathbf{z}}^{(T-1)})$
and $\log q(\mathbf{\tilde{\mathbf{z}}}|\mathbf{x})=\log q^{(0)}(\tilde{\mathbf{z}}^{(0)}|\mathbf{x})+\sum_{t=1}^{T-1}\log q^{(t)}(\tilde{\mathbf{z}}^{(t)}|\tilde{\mathbf{z}}^{(t-1)},\mathbf{x})$.

A stochastic estimation can easily be obtained by replacing the expectation
by an average using a few samples from $q(\widetilde{\mathbf{z}}|\mathbf{x})$.
We can thus compute a lower bound estimate of the average log likelihood
over training, validation and test data. 

Similarly in addition to the lower-bound based on Eq.\ref{eq:lower-bound}
we can use the same few samples from $q(\widetilde{\mathbf{z}}|\mathbf{x})$
to get an importance-sampling estimate of the likelihood based on
Eq. \ref{eq:is-loglikelihood}%
\footnote{Specifically, the two estimates (lower-bound and IS) start by collecting
$k$ samples from $q(\widetilde{\mathbf{z}}|\mathbf{x})$ and computing
for each the corresponding $\ell=\log p(\mathbf{\tilde{\mathbf{z}}},\mathbf{x})-\log q(\widetilde{\mathbf{z}}|\mathbf{x})$.
The lower-bound estimate is then obtained by averaging the resulting
$\ell_{1},\ldots\ell_{k}$, whereas the IS estimate is obtained by
taking the $\log$ of the averaged $e^{\ell_{1}},\ldots,e^{\ell_{k}}$
(in a numerical stable manner as $\mathrm{logsumexp}(\ell_{1},\ldots,\ell_{k})-\log\, k$). %
}.

\subsubsection{Lower-bound-based infusion training procedure}

Since we have derived a lower bound on the likelihood, we can alternatively
choose to optimize this stochastic lower-bound directly during training.
This alternative lower-bound based infusion training procedure differs
only slightly from the denoising-based infusion training procedure
by using $\tilde{\mathbf{z}}^{(t)}$ as a training target at step
$t$ (performing a gradient step to increase $\log p^{(t)}(\tilde{\mathbf{z}}^{(t)}|\tilde{\mathbf{z}}^{(t-1)};\theta^{(t)})$)
whereas denoising training always uses $\mathbf{x}$ as its target
(performing a gradient step to increase $\log p^{(t)}(\mathbf{x}|\tilde{\mathbf{z}}^{(t-1)};\theta^{(t)})$).
Note that the same \emph{reparametrization trick} as used in Variational
Auto-encoders \citep{kingma2014auto} can be used here to backpropagate
through the chain's Gaussian sampling.

\section{Relationship to previously proposed approaches}

\subsection{Markov Chain Monte Carlo for energy-based models}

Generating samples as a repeated application of a Markov transition
operator that operates on input space is at the heart of Markov Chain
Monte Carlo (MCMC) methods. They allow sampling from an energy-model,
where one can efficiently compute the energy or unnormalized negated
log probability (or density) at any point. The transition operator
is then \emph{derived from an explicit energy function} such that
the Markov chain prescribed by a specific MCMC method is guaranteed
to converge to the distribution defined by that energy function, as
the equilibrium distribution of the chain. MCMC techniques have thus
been used to obtain samples from the energy model, in the process
of learning to adjust its parameters.

By contrast here we do not learn an explicit energy function, but
rather learn directly a parameterized transition operator, and define
an \emph{implicit} model distribution based on the result of running
the Markov chain.

\subsection{Variational auto-encoders}

Variational auto-encoders (VAE) \citep{kingma2014auto,DBLP:conf/icml/RezendeMW14}
also start from an unstructured (independent) noise sample and non-linearly
transform this into a distribution that matches the training data.
One difference with our approach is that the VAE typically maps from
a lower-dimensional space to the observation space. By contrast we
learn a stochastic transition operator from input space to input space
that we repeat for $T$ steps. Another key difference, is that the
VAE learns a complex heavily parameterized approximate posterior proposal
$q$ whereas our \emph{infusion based} $q$ can be understood as a
simple heuristic proposal distribution based on $p$. Importantly
the specific heuristic we use to \emph{infuse} $\mathbf{x}$ into
$q$ makes sense precisely because our operator is a map from input
space to input space, and couldn't be readily applied otherwise. The
generative network in \citet{DBLP:conf/icml/RezendeMW14} is a Deep
Latent Gaussian Model (DLGM) just as ours. But their approximate posterior
$q$ is taken to be factorial, including across all layers of the
DLGM, whereas our \emph{infusion based} $q$ involves an ordered sampling
of the layers, as we sample from $q^{(t)}(\tilde{\mathbf{z}}^{(t)}|\tilde{\mathbf{z}}^{(t-1)},\mathbf{x})$. 

More recent proposals involve sophisticated approaches to sample from
better approximate posteriors, as the work of \citet{salimans2015markov}
in which Hamiltonian Monte Carlo is combined with variational inference,
which looks very promising, though computationally expensive, and
\citet{rezende2015variational} that generalizes the use of normalizing
flows to obtain a better approximate posterior.

\subsection{Sampling from autoencoders and Generative Stochastic Networks}

Earlier works that propose to directly learn a transition operator
resulted from research to turn autoencoder variants that have a stochastic
component, in particular denoising autoencoders \citep{vincent2010stacked},
into generative models that one can sample from. This development
is natural, since a stochastic auto-encoder \emph{is} a stochastic
transition operator form input space to input space. Generative Stochastic
Networks (GSN) \citep{alain-al-iai2016-gsns} generalized insights
from earlier stochastic autoencoder sampling heuristics \citep{Rifai-icml2012}
into a more formal and general framework. These previous works on
generative uses of autoencoders and GSNs attempt to learn a chain
whose \emph{equilibrium distribution} will fit the training data.
Because autoencoders and the chain are typically started from or very
close to training data points, they are concerned with the chain mixing
quickly between modes. By contrast our model chain is always restarted
from unstructured noise, and is not required to reach or even have
an equilibrium distribution. Our concern is only what happens during
the $T$ ``burn-in'' initial steps, and to make sure that it transforms
the initial factorial noise distribution into something that best
fits the training data distribution. There are no mixing concerns
beyond those $T$ initial steps. 

A related aspect and limitation of previous denoising autoencoder
and GSN approaches is that these were mainly ``local'' around training
samples: the stochastic operator explored space starting from and
primarily centered around training examples, and learned based on
inputs in these parts of space only. Spurious modes in the generated
samples might result from large unexplored parts of space that one
might encounter while running a long chain.

\subsection{Reversing a diffusion process in non-equilibrium thermodynamics }

The approach of \citet{Sohl-Dickstein-et-al-ICML2015} is probably
the closest to the approach we develop here. Both share a similar
model sampling chain that starts from unstructured factorial noise.
Neither are concerned about an \emph{equilibrium distribution}. They
are however quite different in several key aspects: \citet{Sohl-Dickstein-et-al-ICML2015}
proceed to invert an explicit \emph{diffusion process} that starts
from a training set example and very slowly destroys its structure
to become this random noise, they then learn to reverse this process
i.e. an \emph{inverse diffusion}. To maintain the theoretical argument
that the \emph{exact} reverse process has the same distributional
form (e.g. $p(\mathbf{x}^{(t-1)}|\mathbf{x}^{(t)})$ and $p(\mathbf{x}^{(t)}|\mathbf{x}^{(t-1)})$
both factorial Gaussians), the diffusion has to be infinitesimal by
construction, hence the proposed approaches uses chains with \emph{thousands}
of tiny steps. Instead, our aim is to learn an operator that can yield
a high quality sample efficiently using only a small number $T$ of
larger steps. Also our \emph{infusion} training does not posit a fixed
a priori diffusion process that we would learn to reverse. And while
the distribution of diffusion chain samples of \citet{Sohl-Dickstein-et-al-ICML2015}
is fixed and remains the same all along the training, the distribution
of our infusion chain samples closely follow the model chain as our
model learns. Our proposed infusion sampling technique thus adapts
to the changing generative model distribution as the learning progresses.

Drawing on both \citet{Sohl-Dickstein-et-al-ICML2015} and the walkback
procedure introduced for GSN in \citet{alain-al-iai2016-gsns}, a
variational variant of the walkback algorithm was investigated by
\citet{walkback} at the same time as our work. It can be understood
as a different approach to learning a Markov transition operator,
in which a ``heating'' diffusion operator is seen as a variational
approximate posterior to the forward ``cooling'' sampling operator
with the exact same form and parameters, except for a different temperature.

\section{\label{sec:Experiments} Experiments}

We trained models on several datasets with real-valued examples. We
used as prior distribution $p^{(0)}$ a factorial Gaussian whose parameters
were set to be the mean and variance for each pixel through the training
set. Similarly, our models for the transition operators are factorial
Gaussians. Their mean and elementwise variance is produced as the
output of a neural network that receives the previous $\mathbf{z^{(t-1)}}$
as its input, i.e. $p^{(t)}(\mathbf{z}_{i}^{(t)}|\mathbf{z}^{(t-1)})=\mathcal{N}(\mu_{i}(\mathbf{z}^{(t-1)}),\sigma_{i}^{2}(\mathbf{z}^{(t-1)}))$
where $\mu$ and $\sigma^{2}$ are computed as output vectors of a
neural network. We trained such a model using our\emph{ infusion training}
procedure on MNIST \citep{lecun1998mnist}, Toronto Face Database
\citep{susskind2010toronto}, CIFAR-10 \citep{krizhevsky2009learning},
 and CelebA \citep{liu2015faceattributes}. For all datasets, the
only preprocessing we did was to scale the integer pixel values down
to range {[}0,1{]}. The network trained on MNIST and TFD is a MLP
composed of two fully connected layers with 1200 units using batch-normalization
\citep{ioffe2015batch} %
\footnote{We don't share batch norm parameters across the network, i.e for each
time step we have different parameters and independent batch statistics.%
}. The network trained on CIFAR-10 is based on the same generator as
the GANs of \citet{salimans16-improved}, i.e. one fully connected
layer followed by three transposed convolutions. CelebA was trained
with the previous network where we added another transposed convolution.
We use rectifier linear units \citep{glorot2011deep} on each layer
inside the networks. Each of those networks have two distinct final
layers with a number of units corresponding to the image size. They
use sigmoid outputs, one that predict the mean and the second that
predict a variance scaled by a scalar $\beta$ (In our case we chose
$\beta=0.1$) and we add an epsilon $\epsilon=1e-4$ to avoid an excessively
small variance. For each experiment, we trained the network on 15
steps of denoising with an increasing infusion rate of 1\% ($\omega=0.01,\alpha^{^{(0)}}=0$),
except on CIFAR-10 where we use an increasing infusion rate of 2\%
($\omega=0.02,\alpha^{^{(0)}}=0$) on 20 steps.

\subsection{Numerical results}

\begin{figure*}[htpb]

\begin{minipage}[c][1\totalheight][t]{0.5\columnwidth}%
\begin{figure}[H]
\includegraphics[width=1\columnwidth]{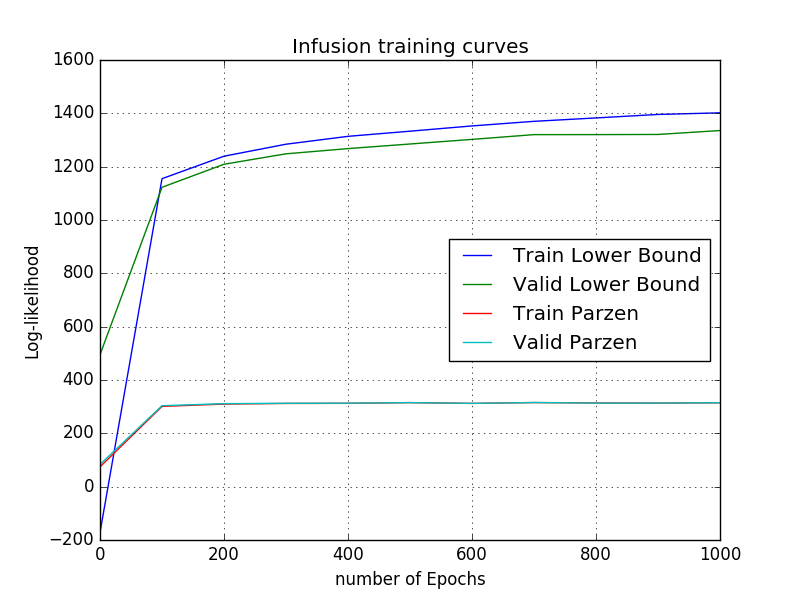}

\protect\caption{\label{fig:Training-curves} Training curves: lower bounds on average
log-likelihood on MNIST as infusion training progresses. We also show
the lower bounds estimated with the Parzen estimation method.}
\end{figure}
\end{minipage}\hfill{} %
\begin{minipage}[c][1\totalheight][t]{0.45\columnwidth}%
\begin{table}[H]
\begin{tabular}{>{\raggedright}p{0.65\columnwidth}|c}
\textbf{Model } & \textbf{Test}\tabularnewline
\hline 
DBM \citep{Bengio-et-al-ICML2013-small}  & $138\pm2$ \tabularnewline
SCAE \citep{Bengio-et-al-ICML2013-small}  & $121\pm1.6$ \tabularnewline
GSN \citep{bengio2014deep}  & $214\pm1.1$ \tabularnewline
Diffusion \citep{Sohl-Dickstein-et-al-ICML2015} & $220\pm1.9$\tabularnewline
GANs (\citeauthor{Goodfellow-et-al-NIPS2014-small})  & $225\pm2$ \tabularnewline
GMMN + AE (\citeauthor{li2015generative}) & $282\pm2$ \tabularnewline
\hline 
Infusion training (Our) & $\mathbf{312}\pm1.7$\tabularnewline
\end{tabular}

\protect\caption{\label{tab:tabParzen}Parzen-window-based estimator of lower bound
on average test log-likelihood on MNIST (in nats).}
\end{table}
\end{minipage}\captionof{table}{\label{tab:tabAIS} Log-likelihood (in nats) estimated by AIS on MNIST test and training sets as reported in \cite{DBLP:journals/corr/WuBSG16} and the log likelihood estimates of our model obtained by infusion training (last three lines). Our initial model uses a Gaussian output with diagonal covariance, and we applied both our lower bound and importance sampling (IS) log-likelihood estimates to it. Since \cite{DBLP:journals/corr/WuBSG16} used only an isotropic output observation model, in order to be comparable to them, we also evaluated our model after replacing the output by an isotropic Gaussian output (same fixed variance for all pixels). Average and standard deviation over 10 repetitions of the evaluation are provided. Note that AIS might provide a higher evaluation of likelihood than our current IS estimate, but this is left for future work.}

\centering%
\begin{tabular}{>{\raggedright}p{0.3\columnwidth}|c|c}
\textbf{Model } & \textbf{Test log-likelihood (1000ex)} & \textbf{Train log-likelihood (100ex)}\tabularnewline
\hline 
VAE-50 (AIS) & $991.435\pm6.477$  & $1272.586\pm6.759$ \tabularnewline
GAN-50 (AIS) & $627.297\pm8.813$ & $620.498\pm31.012$ \tabularnewline
GMMN-50 (AIS) & $593.472\pm8.591$  & $571.803\pm30.864$ \tabularnewline
\hline 
VAE-10 (AIS) & $705.375\pm7.411$  & $780.196\pm19.147$ \tabularnewline
GAN-10 (AIS) & $328.772\pm5.538$ & $318.948\pm22.544$ \tabularnewline
GMMN-10 (AIS) & $346.679\pm5.860$  & $345.176\pm19.893$ \tabularnewline
\hline 
Infusion training + isotropic (IS estimate) & $\mathbf{\mathrm{413.297}}\pm0.460$ & $\mathbf{\mathrm{450.695}}\pm1.617$\tabularnewline
\hline 
Infusion training (IS estimate) & $1836.27\pm0.551$ & $1837.560\pm1.074$\tabularnewline
Infusion training (lower bound) & $1350.598\pm0.079$  & $1230.305\pm0.532$\tabularnewline
\end{tabular}

\end{figure*}

Since we can't compute the exact log-likelihood, the evaluation of
our model is not straightforward. However we use the lower bound estimator
derived in Section \ref{sub:Stochastic-log-likelihood} to evaluate
our model during training and prevent overfitting (see Figure \ref{fig:Training-curves}).
Since most previous published results on non-likelihood based models
(such as GANs) used a Parzen-window-based estimator \citep{breuleux2011quickly},
we use it as our first comparison tool, even if it can be misleading
\citep{Theis2016a}. Results are shown in Table \ref{tab:tabParzen},
we use 10 000 generated samples and $\sigma=0.17$ . To get a better
estimate of the log-likelihood, we then computed both the stochastic
lower bound and the importance sampling estimate (IS) given in Section
\ref{sub:Stochastic-log-likelihood}. For the IS estimate in our MNIST-trained
model, we used 20 000 intermediates samples. In Table \ref{tab:tabAIS}
we compare our model with the recent Annealed Importance Sampling
results \citep{DBLP:journals/corr/WuBSG16}. Note that following their
procedure we add an uniform noise of 1/256 to the (scaled) test point
before evaluation to avoid overevaluating models that might have overfitted
on the 8 bit quantization of pixel values. Another comparison tool
that we used is the Inception score as in \citet{salimans16-improved}
which was developed for natural images and is thus most relevant for
CIFAR-10. Since \citet{salimans16-improved} used a GAN trained in
a semi-supervised way with some tricks, the comparison with our unsupervised
trained model isn't straightforward. However, we can see in Table
\ref{tab:inception} that our model outperforms the traditional GAN
trained without labeled data.

\begin{table}
\protect\caption{\label{tab:inception}Inception score (with standard error) of 50
000 samples generated by models trained on CIFAR-10. We use the models
in \citet{salimans16-improved} as baseline. 'SP' corresponds to the
best model described by \citet{salimans16-improved} trained in a
semi-supervised fashion. '-L' corresponds to the same model after
removing the label in the training process (unsupervised way), '-MBF'
corresponds to a supervised training without minibatch features.}
\center%
\begin{tabular}{l|c|c|c|c|c|}
\textbf{Model } & Real data  & SP  & -L  & -MBF  & Infusion training \tabularnewline
\hline 
\textbf{Inception score} & 11.24 $\pm$ .12  & 8.09 $\pm$ .07  & 4.36 $\pm$ .06  & 3.87 $\pm$ .03  & 4.62 $\pm$ .06 \tabularnewline
\end{tabular}
\end{table}

\subsection{Sample generation}

\begin{figure}
\begin{minipage}[t]{0.49\columnwidth}%
\subfloat[MNIST]{

\includegraphics[width=1\linewidth]{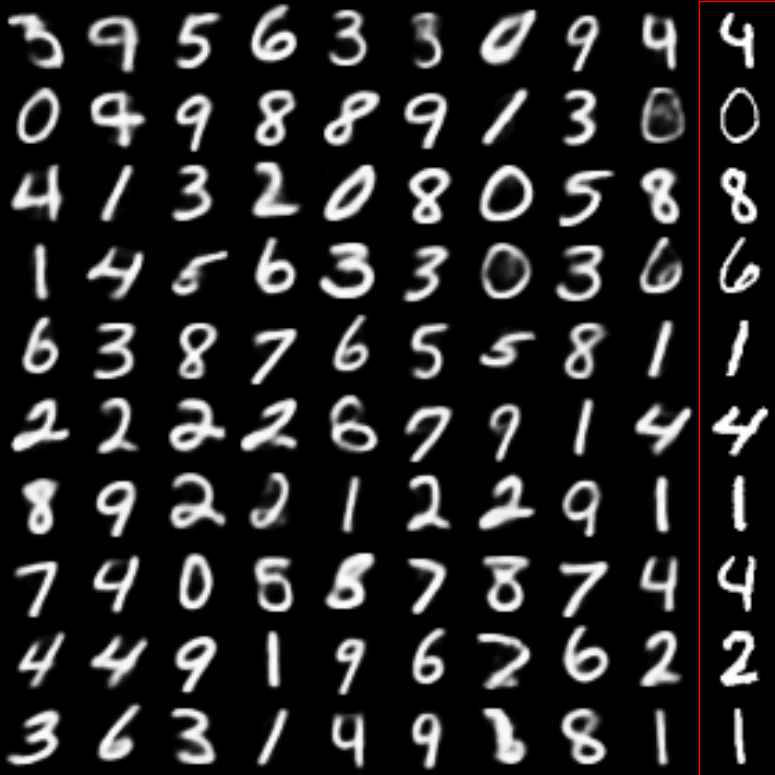}

}%
\end{minipage}\hfill{}%
\begin{minipage}[t]{0.49\columnwidth}%
\subfloat[Toronto Face Dataset]{

\includegraphics[width=1\linewidth]{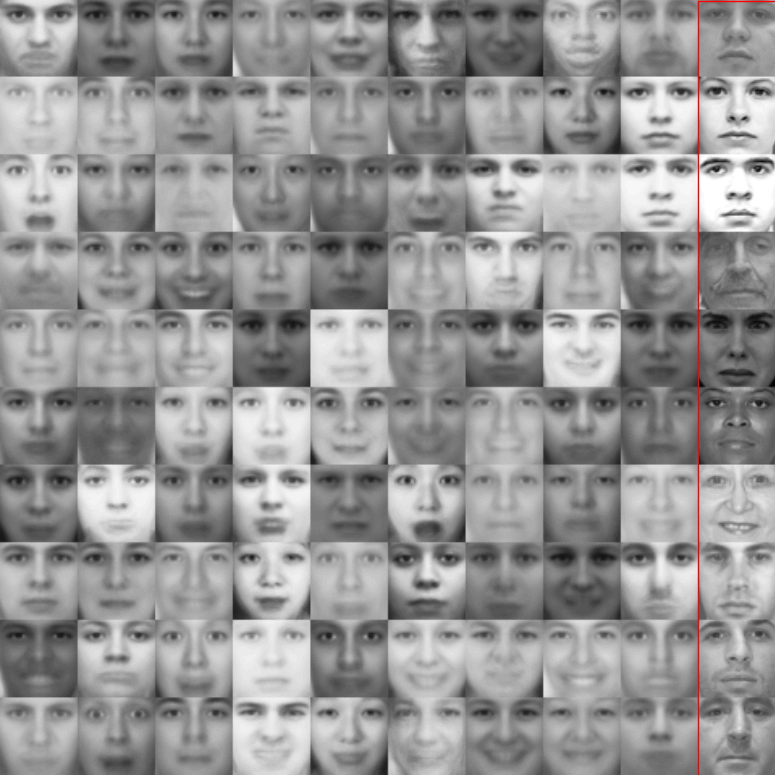}}%
\end{minipage}\\
\\
\begin{minipage}[t]{0.49\columnwidth}%
\subfloat[CIFAR-10]{\includegraphics[width=1\linewidth]{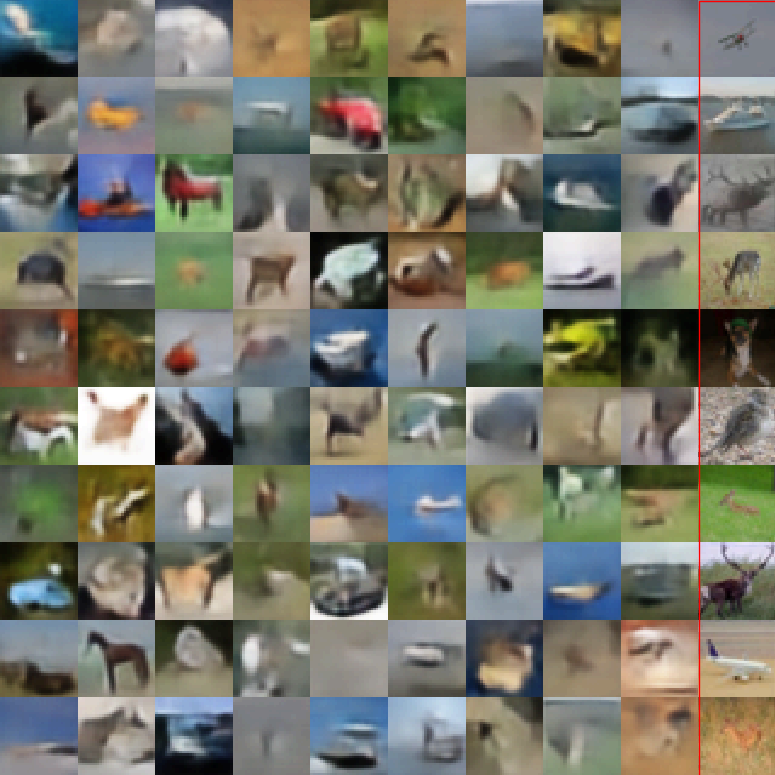}}%
\end{minipage}\hfill{}%
\begin{minipage}[t]{0.49\columnwidth}%
\subfloat[CelebA]{\includegraphics[width=1\linewidth]{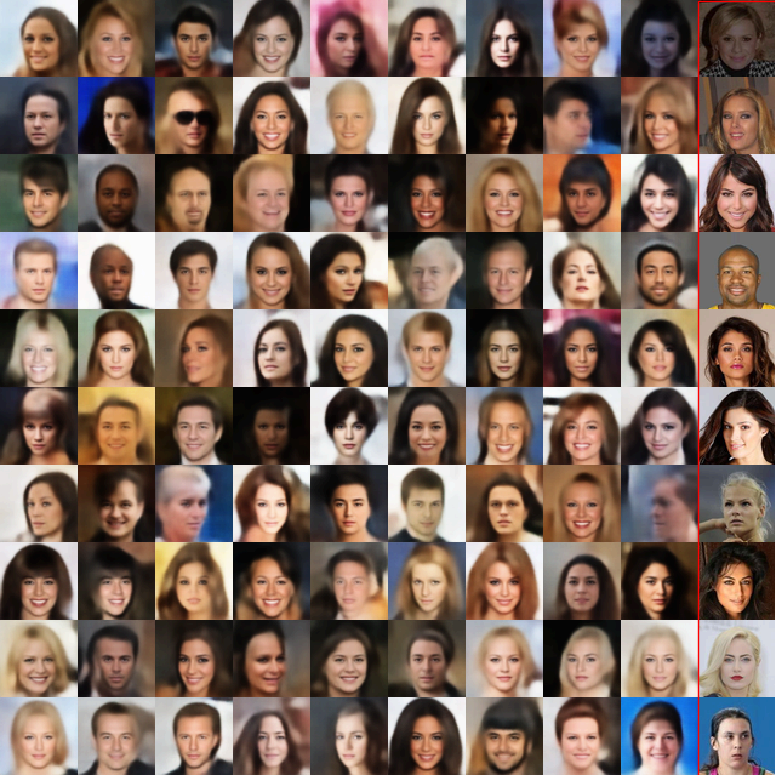}}%
\end{minipage}\protect\caption{\label{fig:samples} Mean predictions by our models on 4 different
datasets. The rightmost column shows the nearest training example
to the samples in the next-to last column.}

\end{figure}

Another common qualitative way to evaluate generative models is to
look at the quality of the samples generated by the model. In Figure
\ref{fig:samples} we show various samples on each of the datasets
we used. In order to get sharper images, we use at sampling time more
denoising steps than in the training time (In the MNIST case we use
30 denoising steps for sampling with a model trained on 15 denoising
steps). To make sure that our network didn't learn to copy the training
set, we show in the last column the nearest training-set neighbor
to the samples in the next-to last column. We can see that our training
method allow to generate very sharp and accurate samples on various
dataset.

\subsection{Inpainting}

\begin{figure}
\center\includegraphics[width=0.8\columnwidth]{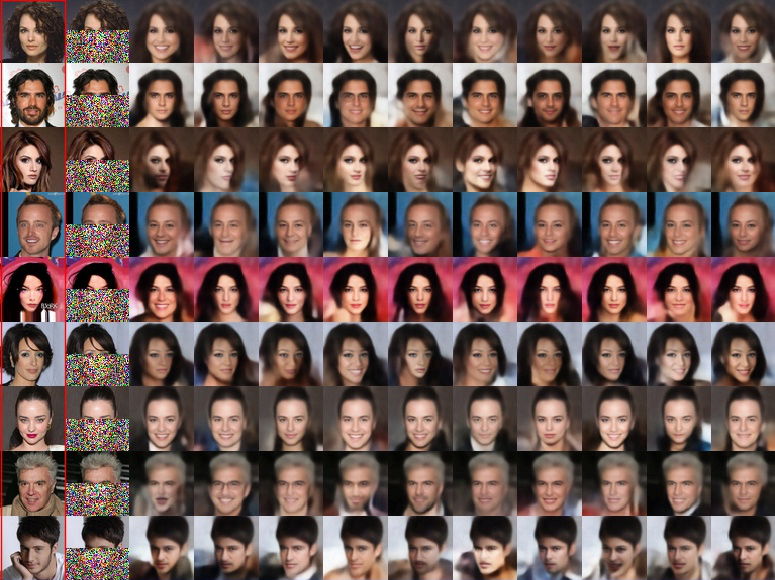}\protect\caption{\label{fig:inpainting}Inpainting on CelebA dataset. In each row,
from left to right: an image form the test set; the same image with
bottom half randomly sampled from our factorial prior. Then several
end samples from our sampling chain in which the top part is clamped.
The generated samples show that our model is able to generate a varied
distribution of coherent face completions. }
\end{figure}

Another method to evaluate a generative model is \emph{inpainting}.
It consists of providing only a partial image from the test set and
letting the model generate the missing part. In one experiment, we
provide only the top half of CelebA test set images and clamp that
top half throughout the sampling chain. We restart sampling from our
model several times, to see the variety in the distribution of the
bottom part it generates. Figure \ref{fig:inpainting} shows that
the model is able to generate a varied set of bottom halves, all consistent
with the same top half, displaying different type of smiles and expression.
We also see that the generated bottom halves transfer some information
about the provided top half of the images (such as pose and more or
less coherent hair cut).

\section{Conclusion and future work}

We presented a new training procedure that allows a neural network
to learn a transition operator of a Markov chain. Compared to the
previously proposed method of \citet{Sohl-Dickstein-et-al-ICML2015}
based on inverting a slow diffusion process, we showed empirically
that infusion training requires far fewer denoising steps, and appears
to provide more accurate models. Currently, many successful generative
models, judged on sample quality, are based on GAN architectures.
However these require to use two different networks, a generator and
a discriminator, whose balance is reputed delicate to adjust, which
can be source of instability during training. Our method avoids this
problem by using only a single network and a simpler training objective. 

Denoising-based infusion training optimizes a heuristic surrogate
loss for which we cannot (yet) provide theoretical guarantees, but
we empirically verified that it results in increasing log-likelihood
estimates. On the other hand the lower-bound-based infusion training
procedure does maximize an explicit variational lower-bound on the
log-likelihood. While we have run most of our experiments with the
former, we obtained similar results on the few problems we tried with
lower-bound-based infusion training.

Future work shall further investigate the relationship and quantify
the compromises achieved with respect to other Markov Chain methods
including \citet{Sohl-Dickstein-et-al-ICML2015,salimans2015markov}
and also to powerful inference methods such as \citet{rezende2015variational}.
As future work, we also plan to investigate the use of more sophisticated
neural net generators, similar to DCGAN's \citep{radford2015unsupervised}
and to extend the approach to a conditional generator applicable to
structured output problems.

\subsubsection*{Acknowledgments}

We would like to thank the developers of Theano \citep{2016arXiv160502688short}
for making this library available to build on, Compute Canada and
Nvidia for their computation resources, NSERC and Ubisoft for their
financial support, and three ICLR anonymous reviewers for helping
us improve our paper.

\bibliographystyle{iclr2017_conference}
\bibliography{infusionNet}

\newpage

\appendix

\section{Details on the experiments}

\subsection{MNIST experiments}

\begin{figure}
\begin{minipage}[t]{0.47\columnwidth}%
\subfloat[Networks trained with 1 infusion step. Each infusion rate in the figure
corresponds to $\alpha^{(0)}$. Since we have only one step, we have
$\omega=0$.]{\includegraphics[width=1\columnwidth]{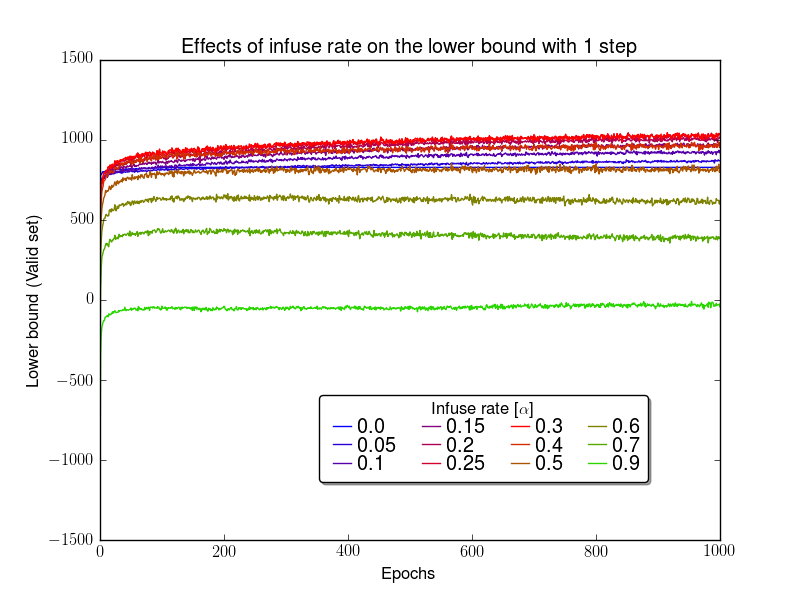}}%
\end{minipage}\hfill{}%
\begin{minipage}[t]{0.47\columnwidth}%
\subfloat[Networks trained with 5 infusion steps. Each infusion rate corresponds
to $\omega$. We set $\alpha^{(0)}=0$.]{\includegraphics[width=1\columnwidth]{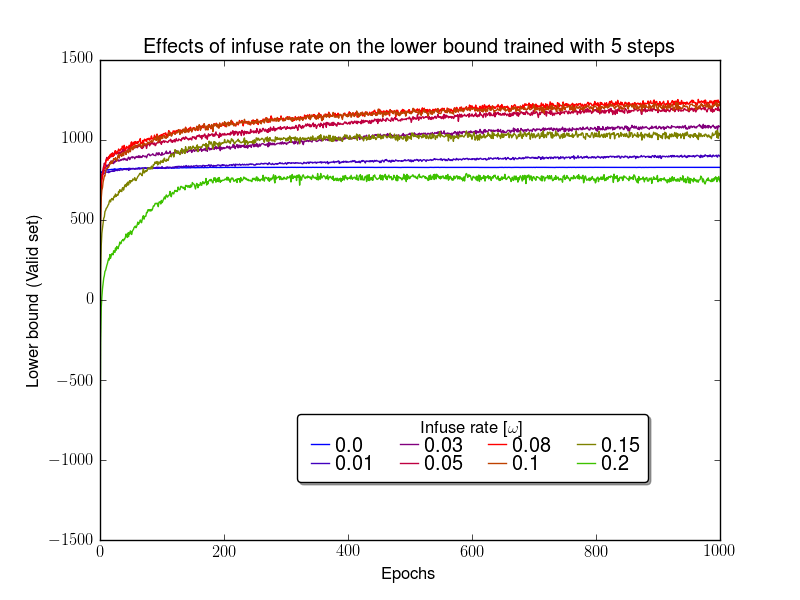}}%
\end{minipage}

\begin{minipage}[t]{0.47\columnwidth}%
\subfloat[Networks trained with 10 infusion steps. Each infusion rate corresponds
to $\omega$. We set $\alpha^{(0)}=0$.]{\includegraphics[width=1\columnwidth]{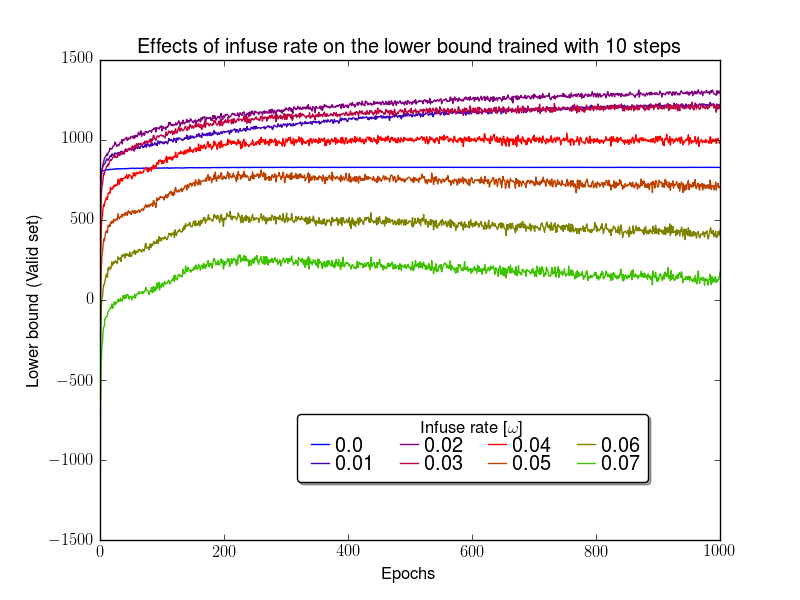}}\hfill{}%
\end{minipage}\hfill{}%
\begin{minipage}[t]{0.47\columnwidth}%
\subfloat[\label{fig:Networks-trained-15}Networks trained with 15 infusion
steps. Each infusion rate corresponds to $\omega$. We set $\alpha^{(0)}=0$.]{\includegraphics[width=1\columnwidth]{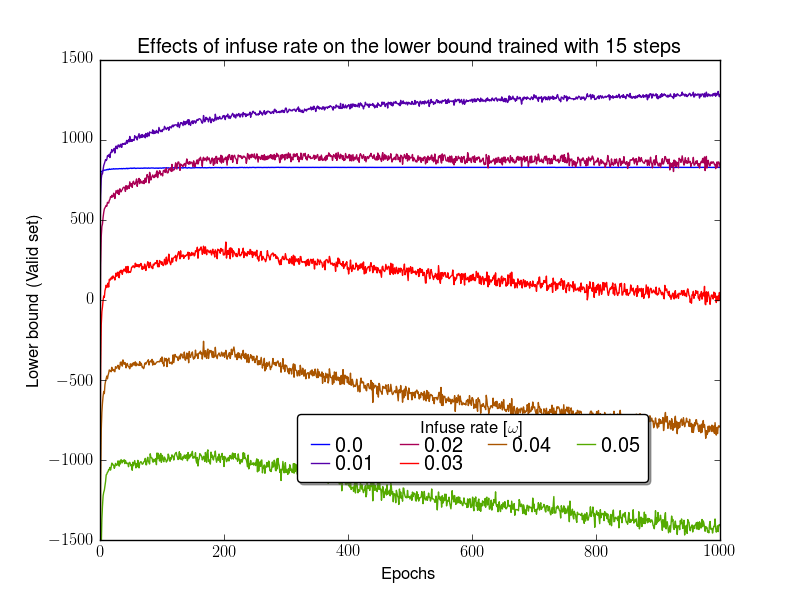}}%
\end{minipage}

\begin{minipage}[t]{0.47\columnwidth}%
\subfloat[\label{fig:Networks-trained-with-10}Networks trained with 10 infusion
steps. In this experiment we use the same infusion rate for each time
step such that $\forall_{t}\alpha^{(t)}=\alpha^{(0)}$. Each infusion
rate in the figure corresponds to different values for $\alpha^{(0)}$.]{\includegraphics[width=1\columnwidth]{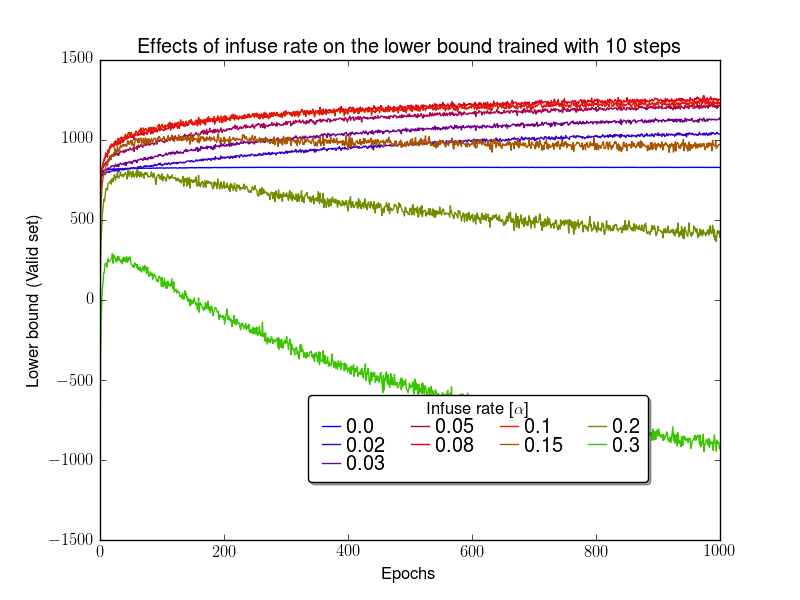}}\hfill{}%
\end{minipage}\hfill{}%
\begin{minipage}[t]{0.47\columnwidth}%
\subfloat[\label{fig:Networks-trained-with-15}Networks trained with 15 infusion
steps. In this experiment we use the same infusion rate for each time
step such that $\forall_{t}\alpha^{(t)}=\alpha^{(0)}$. Each infusion
rate in the figure corresponds to different values $\alpha^{(0)}$.]{\includegraphics[width=1\columnwidth]{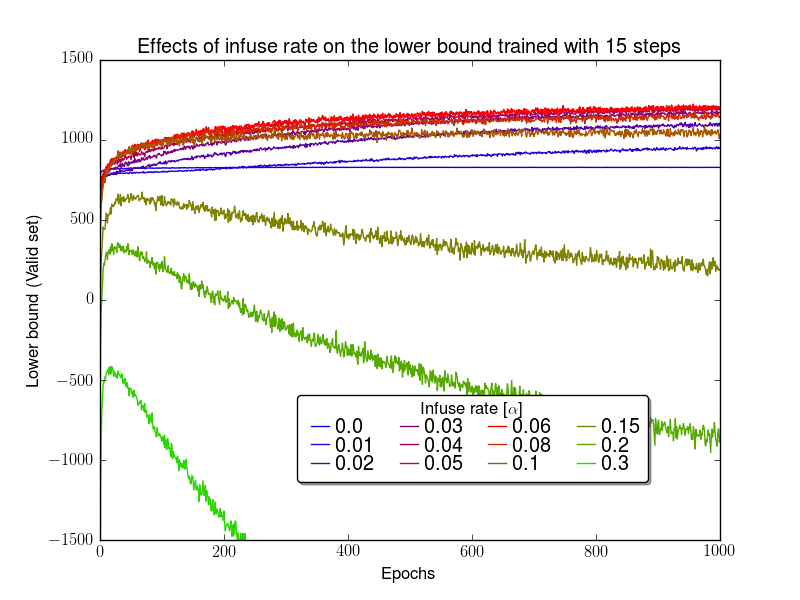}}%
\end{minipage}

\protect\caption{\label{fig:Training-curves-on}Training curves on MNIST showing the
log likelihood lower bound (nats) for different infusion rate schedules
and different number of steps. We use an increasing schedule $\alpha^{(t)}=\alpha^{^{(t-1)}}+\omega$.
In each sub-figure for a fixed number of steps, we show the lower
bound for different infusion rates.}
\end{figure}

We show the impact of the infusion rate $\alpha^{(t)}=\alpha^{^{(t-1)}}+\omega$
for different numbers of training steps on the lower bound estimate
of log-likelihood on the Validation set of MNIST in Figure \ref{fig:Training-curves-on}.
We also show the quality of generated samples and the lower bound
evaluated on the test set in Table \ref{tab:samples_mnist_ll}. Each
experiment in Table \ref{tab:samples_mnist_ll} uses the corresponding
models of Figure \ref{fig:Training-curves-on} that obtained the best
lower bound value on the validation set. We use the same network architecture
as described in Section \ref{sec:Experiments}, i.e two fully connected
layers with Relu activations composed of 1200 units followed by two
distinct fully connected layers composed of 784 units, one that predicts
the means, the other one that predicts the variances. Each mean and
variance is associated with one pixel. All of the the parameters of
the model are shared across different steps except for the batch norm
parameters. During training, we use the batch statistics of the current
mini-batch in order to evaluate our model on the train and validation
sets. At test time (Table \ref{tab:samples_mnist_ll}), we first compute
the batch statistics over the entire train set for each step and then
use the computed statistics to evaluate our model on the test test.

We did some experiments to evaluate the impact of $\alpha$ or $\omega$
in $\alpha^{(t)}=\alpha^{^{(t-1)}}+\omega$. Figure \ref{fig:Training-curves-on}
shows that as the number of steps increases, the optimal value for
infusion rate decreases. Therefore, if we want to use many steps,
we should have a small infusion rate. These conclusions are valid
for both increasing and constant infusion rate. For example, the optimal
$\alpha$ for a constant infusion rate, in Figure \ref{fig:Networks-trained-with-10}
with 10 steps is 0.08 and in Figure \ref{fig:Networks-trained-with-15}
with 15 steps is 0.06. If the number of steps is not enough or the
infusion rate is too small, the network will not be able to learn
the target distribution as shown in the first rows of all subsection
in Table \ref{tab:samples_mnist_ll}. 

In order to show the impact of having a constant versus an increasing
infusion rate, we show in Figure \ref{fig:In-those-figures,} the
samples created by infused and sampling chains. We observe that having
a small infusion rate over many steps ensures a slow blending of the
model distribution into the target distribution.

In Table 4, we can see high lower bound values on the test set with
few steps even if the model can\textquoteright t generate samples
that are qualitatively satisfying. These results indicate that we
can\textquoteright t rely on the lower bound as the only evaluation
metric and this metric alone does not necessarily indicate the suitability
of our model to generated good samples. However, it is still a useful
tool to prevent overfitting (the networks in Figure \ref{fig:Networks-trained-with-10}
and \ref{fig:Networks-trained-with-15} overfit when the infusion
rate becomes too high). Concerning the samples quality, we observe
that having a small infusion rate over an adequate number of steps
leads to better samples.

\begin{table}
\protect\caption{\label{tab:samples_mnist_ll} Infusion rate impact on the lower bound
log-likelihood (test set) and the samples generated by a network trained
with different number of steps. Each sub-table corresponds to a fixed
number of steps. Each row corresponds to a different infusion rate,
where we show its lower bound and also its corresponding generated
samples from the trained model. Note that for images, we show the
mean of the Gaussian distributions instead of the true samples. As
the number of steps increases, the optimal infusion rate decreases.
Higher number of steps contributes to better qualitative samples,
as the best samples can be seen with 15 steps using $\alpha=0.01$.}

\subfloat[infusion rate impact on the lower bound log-likelihood (test set)
and the samples generated by a network trained with 1 step.]{%
\begin{tabular}[t]{|c|c|l|}
\hline 
infusion rate & Lower bound (test) & Means of the model\tabularnewline
\hline 
0.0 & 824.34 & \multirow{12}{*}{\includegraphics[width=0.6\columnwidth,height=0.33\columnwidth]{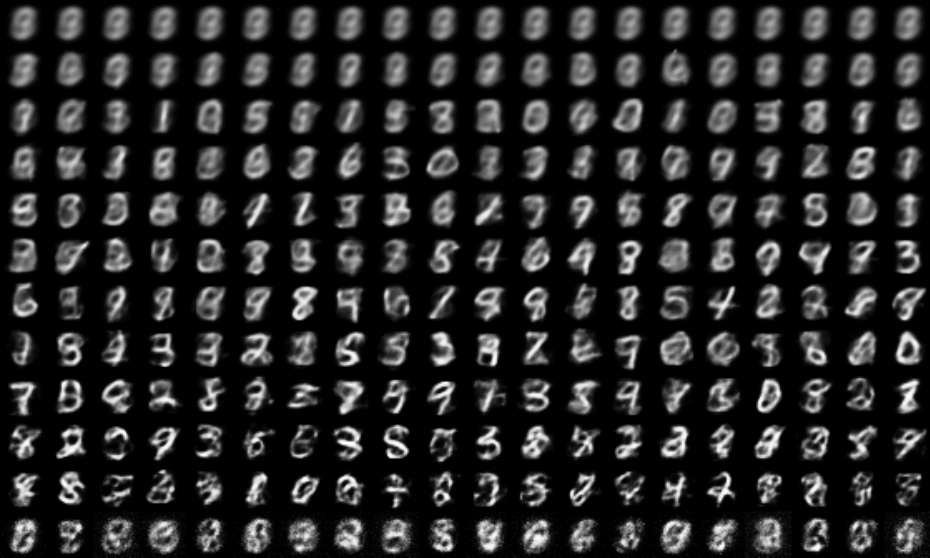}}\tabularnewline
\cline{1-2} 
0.05 & 885.35 & \tabularnewline
\cline{1-2} 
0.1 & 967.25 & \tabularnewline
\cline{1-2} 
0.15 & 1063.27 & \tabularnewline
\cline{1-2} 
0.2 & 1115.15 & \tabularnewline
\cline{1-2} 
0.25 & 1158.81 & \tabularnewline
\cline{1-2} 
$\mathbf{0.3}$ & $\mathbf{1209.39}$ & \tabularnewline
\cline{1-2} 
0.4 & 1209.16 & \tabularnewline
\cline{1-2} 
0.5 & 1132.05 & \tabularnewline
\cline{1-2} 
0.6 & 1008.60 & \tabularnewline
\cline{1-2} 
0.7 & 854.40 & \tabularnewline
\cline{1-2} 
0.9 & -161.37 & \tabularnewline
\hline 
\end{tabular}}

\subfloat[\label{tab:infusion-rate-1}infusion rate impact on the lower bound
log-likelihood (test set) and the samples generated by a network trained
with 5 steps]{%
\begin{tabular}[t]{|c|c|l|}
\hline 
infusion rate & Lower bound (test) & \tabularnewline
\hline 
0.0 & 823.81 & \multirow{8}{*}{\includegraphics[width=0.6\columnwidth,height=0.22\columnwidth]{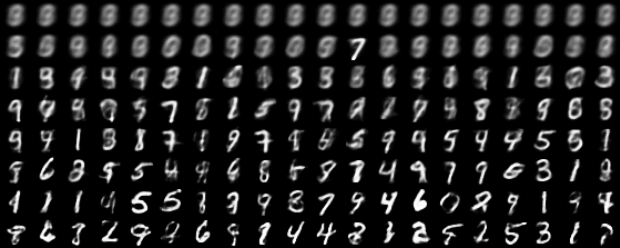}}\tabularnewline
\cline{1-2} 
0.01 & 910.19 & \tabularnewline
\cline{1-2} 
0.03 & 1142.43 & \tabularnewline
\cline{1-2} 
0.05 & 1303.19 & \tabularnewline
\cline{1-2} 
0.08 & 1406.38 & \tabularnewline
\cline{1-2} 
$\mathbf{0.1}$ & $\mathbf{1448.66}$ & \tabularnewline
\cline{1-2} 
0.15 & 1397.41 & \tabularnewline
\cline{1-2} 
0.2 & 1262.57 & \tabularnewline
\hline 
\end{tabular}}

\subfloat[\label{tab:infusion-rate-5}infusion rate impact on the lower bound
log-likelihood (test set) and the samples generated by a network trained
with 10 steps]{%
\begin{tabular}[t]{|c|c|l|}
\hline 
infusion rate & Lower bound (test) & \tabularnewline
\hline 
0.0 & 824.42 & \multirow{8}{*}{\includegraphics[width=0.6\columnwidth,height=0.22\columnwidth]{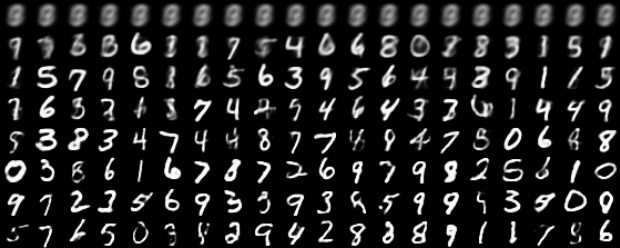}}\tabularnewline
\cline{1-2} 
0.01 & 1254.07 & \tabularnewline
\cline{1-2} 
$\mathbf{0.02}$ & $\mathbf{1389.12}$ & \tabularnewline
\cline{1-2} 
0.03 & $1366.6$8 & \tabularnewline
\cline{1-2} 
0.04 & 1223.47 & \tabularnewline
\cline{1-2} 
0.05 & 1057.43 & \tabularnewline
\cline{1-2} 
0.05 & 846.73 & \tabularnewline
\cline{1-2} 
0.07 & 658.66 & \tabularnewline
\hline 
\end{tabular}}

\subfloat[infusion rate impact on the lower bound log-likelihood (test set)
and the samples generated by a network trained with 15 steps]{%
\begin{tabular}[t]{|c|c|l|}
\hline 
infusion rate & Lower bound (test) & \tabularnewline
\hline 
0.0 & 824.50 & \multirow{7}{*}{\includegraphics[width=0.6\columnwidth,height=0.19\columnwidth]{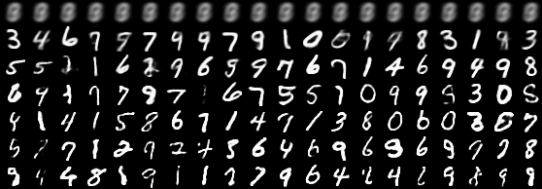}}\tabularnewline
\cline{1-2} 
$\mathbf{0.01}$ & $\mathbf{1351.03}$ & \tabularnewline
\cline{1-2} 
0.02 & 1066.60 & \tabularnewline
\cline{1-2} 
0.03 & 609.10 & \tabularnewline
\cline{1-2} 
0.04 & 876.93 & \tabularnewline
\cline{1-2} 
0.05 & -479.69 & \tabularnewline
\cline{1-2} 
0.06 & -941.78 & \tabularnewline
\hline 
\end{tabular}}
\end{table}

\begin{figure}
\begin{minipage}[t]{0.45\columnwidth}%
\subfloat[\label{fig:Chain-infused_cte}Chains infused with MNIST test set samples
by a constant rate ($\alpha^{(0)}=0.05,\enskip\omega=0$) in 15 steps.]{\includegraphics[width=1\columnwidth]{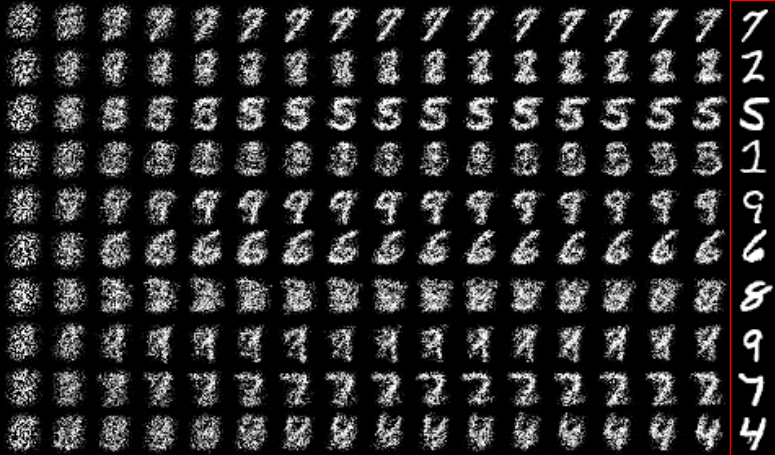}}%
\end{minipage}\hfill{}%
\begin{minipage}[t]{0.45\columnwidth}%
\subfloat[\label{fig:Generating-chain-cte}Model sampling chains on MNIST using
a network trained with a constant infusion rate ($\alpha^{(0)}=0.05,\enskip\omega=0$)
in 15 steps.]{\includegraphics[width=1\columnwidth,height=0.6\columnwidth]{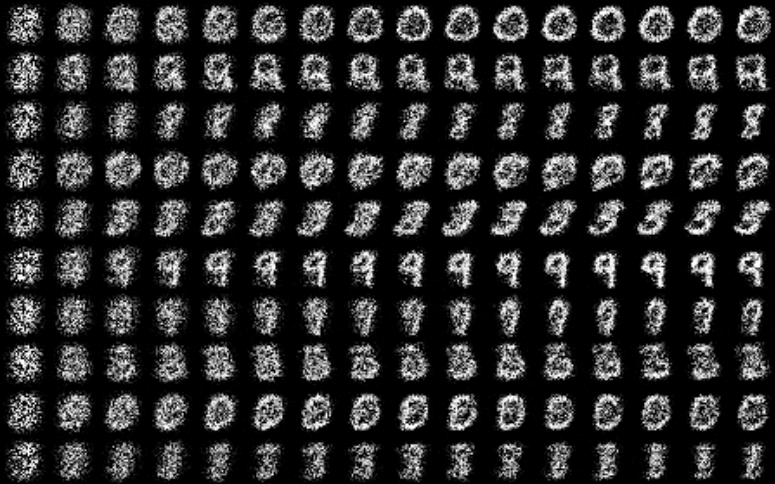}}%
\end{minipage}

\begin{minipage}[t]{0.45\columnwidth}%
\subfloat[\label{fig:Chain-infused-in}Chains infused with MNIST test set samples
by an increasing rate ($\alpha^{(0)}=0.0,\enskip\omega=0.01$) in
15 steps.]{\includegraphics[width=1\columnwidth]{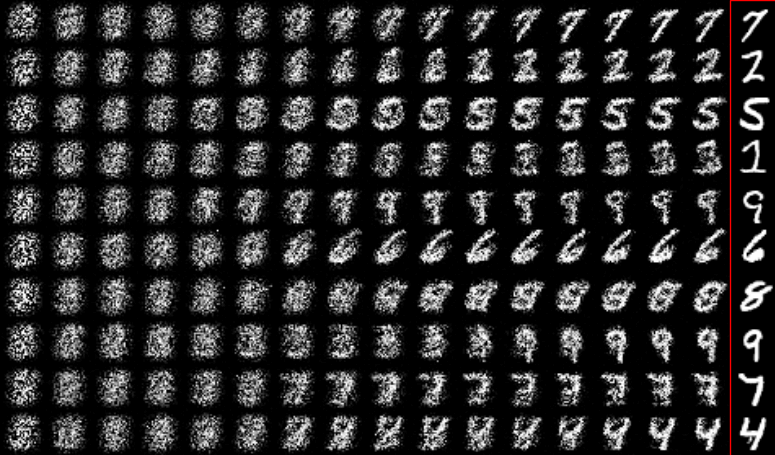}}%
\end{minipage}\hfill{}%
\begin{minipage}[t]{0.45\columnwidth}%
\subfloat[\label{fig:Generating-chain-in}Model sampling chains on MNIST using
a network trained with an increasing infusion rate ($\alpha^{(0)}=0.0,\enskip\omega=0.01$)
in 15 steps.]{\includegraphics[width=1\columnwidth,height=0.6\columnwidth]{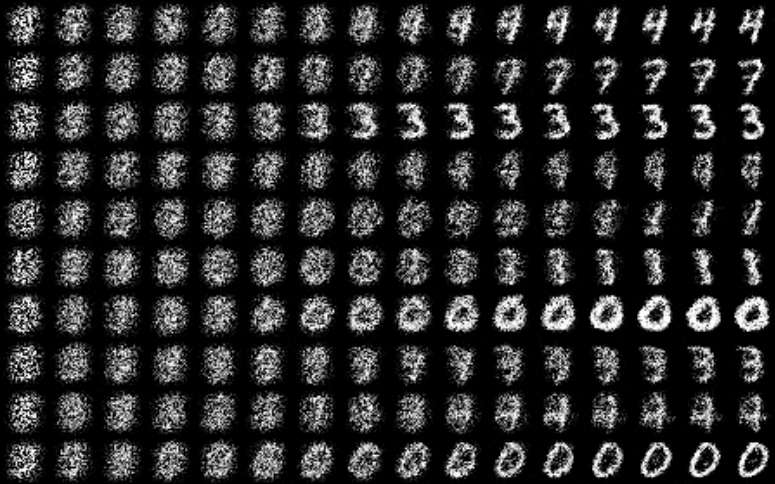}}%
\end{minipage}

\protect\caption{\label{fig:In-those-figures,}Comparing samples of constant infusion
rate versus an increasing infusion rate on infused and generated chains.
The models are trained on MNIST in 15 steps. Note that having an increasing
infusion rate with a small value for $\omega$ allows a slow convergence
to the target distribution. In contrast having a constant infusion
rate leads to a fast convergence to a specific point. Increasing infusion
rate leads to more visually appealing samples. We observe that having
an increasing infusion rate over many steps ensures a slow blending
of the model distribution into the target distribution.}
\end{figure}

\subsection{Infusion and model sampling chains on natural images datasets}

\begin{figure}
\subfloat[\label{fig:Infusion-chains-on}Infusion chains on CIFAR-10. Last column
corresponds to the target used to infuse the chain.]{\includegraphics[scale=0.7]{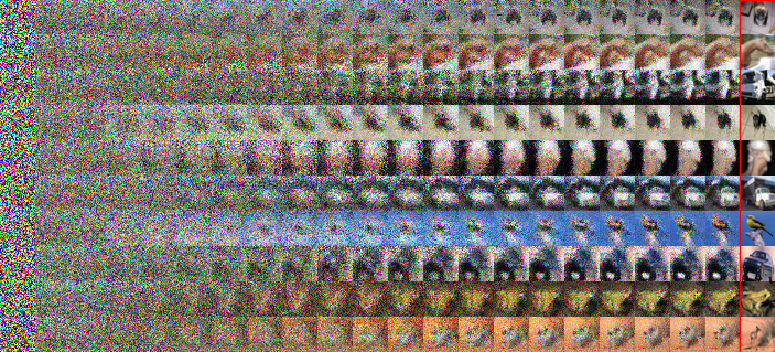}

}

\subfloat[\label{fig:Generating-chains-on}Model sampling chains on CIFAR-10]{\includegraphics[scale=0.7]{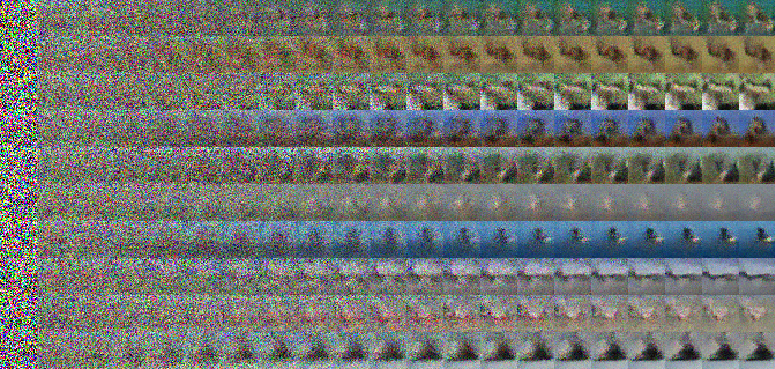}

}

\protect\caption{\label{fig:Chains-on-cifar10.}Infusion chains (Sub-Figure \ref{fig:Infusion-chains-on})
and model sampling chains (Sub-Figure \ref{fig:Generating-chains-on})
on CIFAR-10.}
\end{figure}

\begin{figure}
\subfloat[Infusion chains on CelebA. Last column corresponds to the target used
to infuse the chain.]{\includegraphics[scale=0.7]{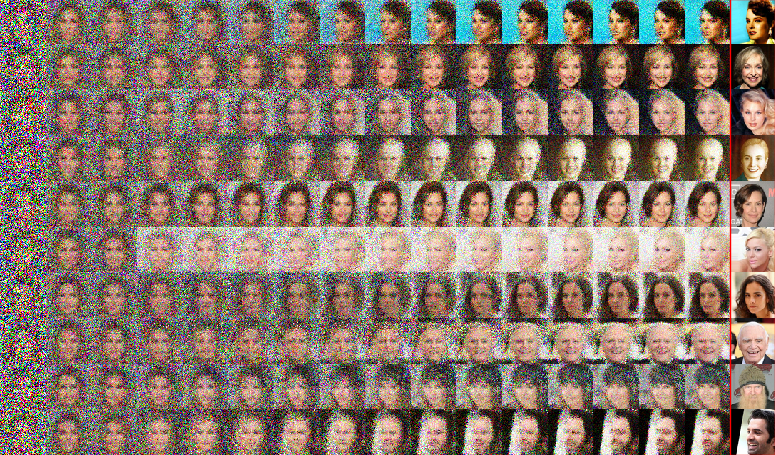}\label{fig:celebA:infuse}}

\subfloat[Model sampling chains on CelebA]{\includegraphics[scale=0.7]{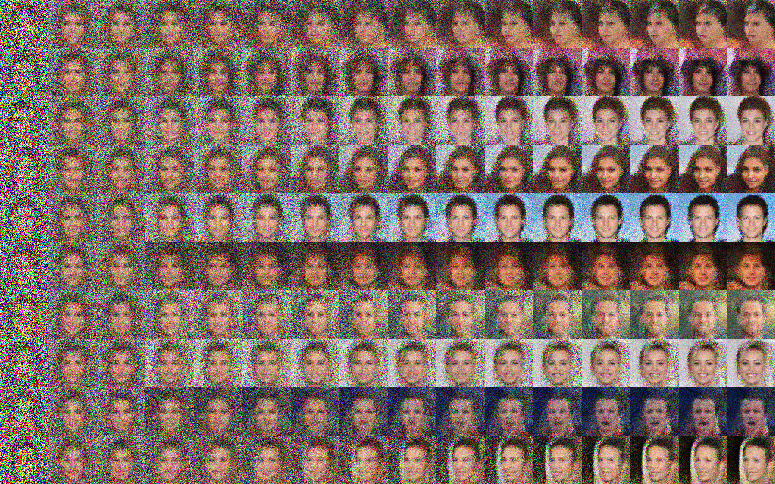}\label{fig:celebA:gen}}

\protect\caption{\label{fig:Chains-on-celeba}Infusion chains (Sub-Figure \ref{fig:celebA:infuse})
and model sampling chains (Sub-Figure \ref{fig:celebA:gen}) on CelebA.}
\end{figure}

In order to show the behavior of our model trained by Infusion on
more complex datasets, we show in Figure \ref{fig:Chains-on-cifar10.}
chains on CIFAR-10 dataset and in Figure \ref{fig:Chains-on-celeba}
chains on CelebA dataset. In each Figure, the first sub-figure shows
the chains infused by some test examples and the second sub-figure
shows the model sampling chains. In the experiment on CIFAR-10, we
use an increasing schedule $\alpha^{(t)}=\alpha^{^{(t-1)}}+0.02$
with $\alpha^{(0)}=0$ and 20 infusion steps (this corresponds to
the training parameters). In the experiment on CelebA, we use an increasing
schedule $\alpha^{(t)}=\alpha^{^{(t-1)}}+0.01$ with $\alpha^{(0)}=0$
and 15 infusion steps.

\section*{}
\end{document}